\documentclass{bmvc2k}
\usepackage{adjustbox}
\usepackage{enumitem}
\usepackage{booktabs}
\usepackage{pgfplots}

\pgfplotsset{compat=1.9}

\definecolor{RYB1}{RGB}{141, 211, 199}
\definecolor{RYB2}{RGB}{255, 255, 179}
\definecolor{RYB3}{RGB}{190, 186, 218}
\definecolor{RYB4}{RGB}{251, 128, 114}
\definecolor{RYB5}{RGB}{128, 177, 211}
\definecolor{RYB6}{RGB}{253, 180, 98}
\definecolor{RYB7}{RGB}{179, 222, 105}

\usepackage{floatrow}
\newfloatcommand{capbtabbox}{table}[][\FBwidth]
\usepackage{stackengine}

\title{DublinCity: Annotated LiDAR Point Cloud and its Applications} 

\addauthor{S. M. Iman Zolanvari}{zolanvari.com}{1}
\addauthor{Susana Ruano}{ruanosas@scss.tcd.ie}{1}
\addauthor{Aakanksha Rana}{ranaa@scss.tcd.ie}{1}
\addauthor{Alan Cummins}{alan.cummins@tcd.ie}{1}
\addauthor{Rogerio Eduardo da Silva}{www.rogerioesilva.net}{2}
\addauthor{Morteza Rahbar}{rahbar@arch.ethz.ch}{34}
\addauthor{Aljosa Smolic}{smolica@scss.tcd.ie}{1}
\addinstitution{
 V-SENSE\\
 School of Computer Science and Statistics \\
 Trinity College Dublin, Ireland\\
}

\addinstitution{
 University of Houston-Victoria,
 Victoria, Texas, US\\
}
\addinstitution{
 CAAD\\
 Department of Architecture \\
 ETH, Zurich, Switzerland
}
\addinstitution{
 Department of Architecture\\
 Tarbiat Modares University\\
 Tehran, Iran
}

\runninghead{DublinCity}{Annotated LiDAR Point Cloud and its Applications}

\def\eg{\emph{e.g.}~}
\def\etal{\emph{et al}\bmvaOneDot}
\def\ie{\emph{i.e.}~}
\begin{document}
\maketitle
\begin{abstract}
Scene understanding of full-scale 3D models of an urban area remains a challenging task. While advanced computer vision techniques offer cost-effective approaches to analyse 3D urban elements, a precise and densely labelled dataset is quintessential. The paper presents the first-ever labelled dataset for a highly dense Aerial Laser Scanning (ALS) point cloud at city-scale. This work introduces a novel benchmark dataset that includes a manually annotated point cloud for over 260 million laser scanning points into 100'000 (approx.) assets from Dublin LiDAR point cloud \cite{laefer20172015} in 2015. Objects are labelled into 13 classes using hierarchical levels of detail from large (\ie building, vegetation and ground) to refined (\ie window, door and tree) elements. To validate the performance of our dataset, two different applications are showcased. Firstly, the labelled point cloud is employed for training Convolutional Neural Networks (CNNs) to classify urban elements. The dataset is tested on the well-known state-of-the-art CNNs (\ie PointNet, PointNet++ and So-Net). Secondly, the complete ALS dataset is applied as detailed ground truth for city-scale image-based 3D reconstruction.
\end{abstract}
%
\section{Introduction}
In computer vision, automated identification of three-dimensional (3D) assets in an unstructured large dataset is essential for scene understanding. In order to collect such a big dataset at city-scale, laser scanning technology, also known as Light Detection and Ranging (LiDAR), offers an efficient means of capture. Aerial Laser Scanning (ALS), a LiDAR data acquisition approach, is generally used for obtaining data for a large area (\eg an entire urban region). The data consists of unprocessed waveforms and a discrete point cloud. In addition, image data can be collected alongside this LiDAR point cloud data if an additional camera is used.
 
To develop algorithms which are able to automatically identify 3D shapes and objects in a large LiDAR dataset, it is crucial to have a well-annotated ground-truth data available. However, the generation of such labels is cumbersome and expensive. It is essential to have access to a full 3D, dense, and non-synthetic labelled point cloud at city-scale that includes a variety of urban elements (\eg various types of roofs, buildings' facade, windows, trees and sidewalks). While there have been several attempts to generate such a labelled dataset by using photogrammetric or morphological methods, review of the well-known available datasets shows that none of these can completely satisfy all requirements \cite{meng2010ground}. Therefore, this paper presents a novel manually annotated point cloud from ALS data of Dublin city centre that was obtained by Laefer \etal.~\cite{laefer20172015}. This dataset is one of the densest urban aerial LiDAR datasets that have ever been collected with an average point density of $348.43$ points/m$^2$~\cite{gupta2018automatic}.

The main contribution of this paper is the manual annotation of over 260 million laser scanning points into 100'000 (approx.) assets into 13 classes (\eg building, tree, facade, windows and streets) with the hierarchical levels. The proposed labelled dataset is the first of its kind regarding its accuracy, density and diverse classes, particularly with the city scale coverage area. To the best knowledge of the authors, no publicly available LiDAR dataset is available with the unique features of the DublinCity dataset. The hierarchical labels offer excellent potential for classification and semantic segmentation of urban elements from refined (\eg windows and doors) to coarse level (\eg buildings and ground). Herein, two different applications are introduced to validate the important usage of the labelled point cloud. 

The first application is an automated technique for classification of 3D objects by using three state-of-the-art CNN based models. Machine Learning (ML) techniques offer relatively efficient and high accuracy means to process massive datasets~\cite{wu20153d}. ML techniques highly rely on input training datasets. Most datasets are not generated for detailed 3D urban elements on a city-scale. Therefore, there is limited research in that direction. Herein, three highly cited CNNs are trained and evaluated by employing the introduced manually labelled point cloud. Secondly, in the original dataset, aerial images were also captured during the helicopter fly-over. By using this image data, an image-based 3D reconstruction of the city is generated and the result is compared to the LiDAR data across multiple factors. The next section reviews previous related datasets and the aforementioned relevant applications in more detail.
%
\section{Related Work}
\label{sec:RelatedWork}

Several state-of-the-art annotated point cloud datasets have provided significant opportunities to develop and improve 3D analysis algorithms for various related research fields (\eg computer vision, remote sensing, autonomous navigation and urban planning). However, due to varying limitations, such datasets are not ideal to achieve the necessary accuracy. Hence, the availability of high quality and large size point cloud datasets is of utmost importance.

In the ShapeNet dataset~\cite{wu20153d}, point clouds are not obtained from scanning of real objects, rather they are generated from 3D synthetic CAD models. In TerraMobilita/iQmulus~\cite{vallet2015terramobilita} and more recently in the Paris-Lille~\cite{roynard2018paris} project, the datasets are obtained by Mobile Laser Scanning (MLS). However, MLS LiDAR datasets are not fully 3D, as MLS datasets can only scan the ground and buildings' facades without any information from the roof or the other sides of buildings. While the ScanNet~\cite{dai2017scannet} dataset consists of real 3D objects, it only includes indoor objects (\eg chairs, desks and beds) without any elements from outside (\eg buildings' facade or roofs). In contrast, the Semantic3D~\cite{hackel2017semantic3d} dataset generated outdoor point cloud from several registered Terrestrial Laser Scanning (TLS). However, it only covers a small portion of a city with a limited number of elements. Similarly, RoofN3D~\cite{wichmann2018roofn3d} only includes a specific type of element (\ie roofs) without even coverage of distinct urban roofs (\eg flat, pyramid, shed and M-shaped). More recently, the TorontoCity~\cite{wang2017torontocity} dataset was introduced which uses high-precision maps from multiple sources to create the ground truth labels. However, the dataset still has no manually labelled objects. In addition to these datasets, AHN 1,2 and 3 datasets \cite{AHN} also provided large scale LiDAR data for a vast area of Netherlands. While the AHN datasets covered a large area, the average density of the point cloud is only around 8 to 60 points/m${^2}$ which is not sufficient for generation of a detailed 3D model.  

 

\paragraph{3D Point Cloud Object Classification.}
Classification using the 3D point cloud data has gained considerable attention across several research domains \eg autonomous navigation \cite{Qi_2018_CVPR,rana}, virtual and augmented reality creation \cite{hololens,icassp2019} and urban\cite{lang201813, tarabalka}-forest monitoring \cite{pirotti} tasks. Amongst the state-of-the-art classification techniques, CNN based models offer a reliable and cost-effective solution to process 3D point cloud datasets that are massive and unstructured in nature. The earliest CNN model was introduced by~\cite{voxnet}, where voxel grids are used as inputs. 
With rapid advancement in deep learning models, recently, several methods ~\cite{qi2017pointnet,qi2017pointnetplusplus,li2018sonet} have been proposed in the literature that utilise the point cloud data to train the CNN rather than the voxel grids~\cite{voxelnet,voxnet} or collections of images.

PointNet~\cite{qi2017pointnet,qi2017pointnetplusplus} is one of the first successful attempts to describe the point cloud object using a distinctive global feature representation. As an extension to the former, PointNet++~\cite{qi2017pointnetplusplus} is designed to further address the `fine' local details in objects by building a pyramid-like feature aggregation model. To achieve better performance, the recently proposed SO-Net model in~\cite{li2018sonet} formulates the single feature vector representation using the hierarchical feature extraction on individual points. 

\label{subsec:RW:3DRecons}

\paragraph{Image-based 3D Reconstruction.}
Recently, driven by the collection of large imagery datasets, the classic computer vision challenge of image-based 3D reconstruction has been revisited to focus on large-scale problems~\cite{bodis2014fast,schoenberger2016sfm}. As a result, the list of open-source methods for Structure-from-Motion (SfM)~\cite{snavely2008modeling,sweeney2015theia,moulon2013globalOPENMVG,schoenberger2016sfm} and Multi-view Stereo (MVS)~\cite{furukawa2010accurate,barnes2009patchmatch,schoenberger2016mvs} has increased. However, one of the main issues working with these algorithms is the lack of ground truth for measuring the accuracy of the reconstruction, which is usually acquired with active techniques (\eg LiDAR) and, therefore, it is rarely available for large-scale areas~\cite{knapitsch2017tanks,marelli2018blender}. %

There are several benchmarks available for 3D reconstruction. One of the first is the Middlebury benchmark~\cite{seitz2006comparison} which has been recently extended~\cite{aanaes2016large}. In~\cite{aanaes2016large} database, the number of images and objects has increased. However, they provide less than 100 images for each reconstructed model and focus on small objects in a confined indoor space. The EPFL benchmark~\cite{strecha2008benchmarking} and the ETH3D benchmark~\cite{schops2017multi} are also presented to fill the gap between image-based reconstruction techniques and the LiDAR outdoors. While the benchmark dataset is acquired with terrestrial LiDAR, the number of images and models are limited and they only focus on a few monuments. A dataset that overcomes the limitation of the terrestrial LiDAR is the Toronto/Vaihingen ISPRS used in~\cite{zhang2018large}, but it still covers a small area of the city compared to the aerial LiDAR dataset available for the city of Dublin which is presented in this work. In most recent benchmarks~\cite{schops2017multi,knapitsch2017tanks,marelli2018blender} COLMAP~\cite{schoenberger2016sfm,schoenberger2016mvs} is reported as the best approach (SfM + MVS) to generate the reconstructions in the majority of the scenarios. Therefore, in this work we use it to generate the  image-based reconstructions.


%
\section{Database Specification}
\label{sec:Database}
The initial dataset \cite{laefer20172015} includes a major area of Dublin city centre (\ie around 5.6 km${^2}$ including partially covered areas) was scanned via an ALS device which was carried out by helicopter in 2015. However, the actual focused area was around 2 km${^2}$ which contains the most densest LiDAR point cloud and imagery dataset. The flight altitude was mostly around 300m and the total journey was performed in 41 flight path strips. 

\begin{figure}
\footnotesize
\centering
 \resizebox{.9\textwidth}{!}{
 \def\arraystretch{1.5}
 \begin{tabular}{p{0.45\textwidth}p{0.45\textwidth}}
{\includegraphics[width=0.46\textwidth]{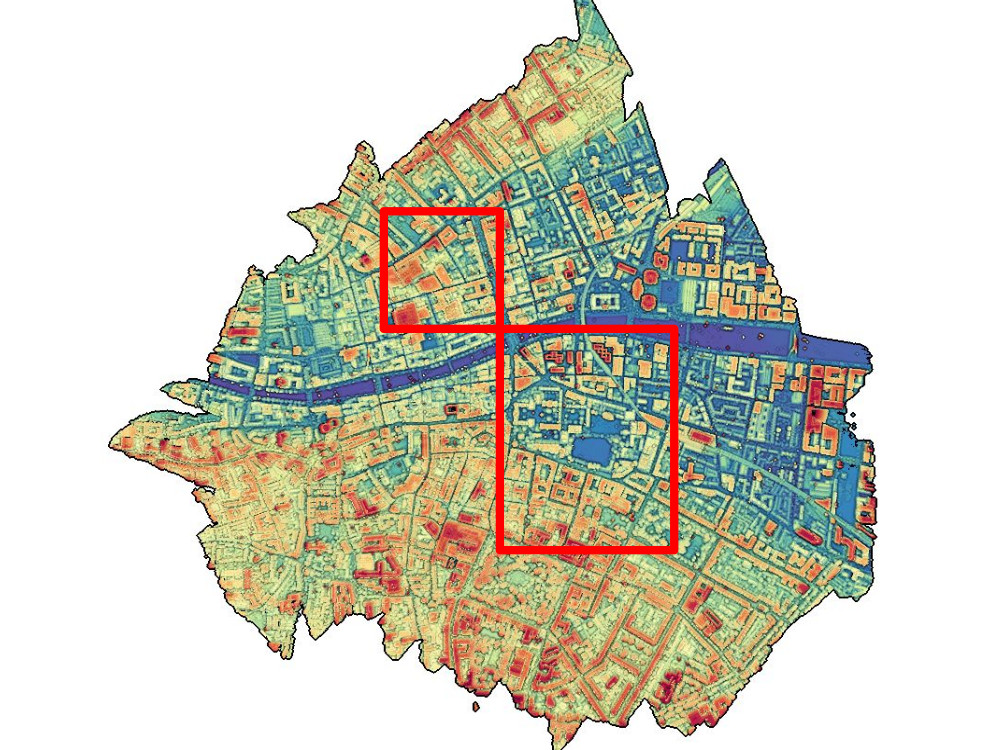}}&
{\includegraphics[width=.51\textwidth]{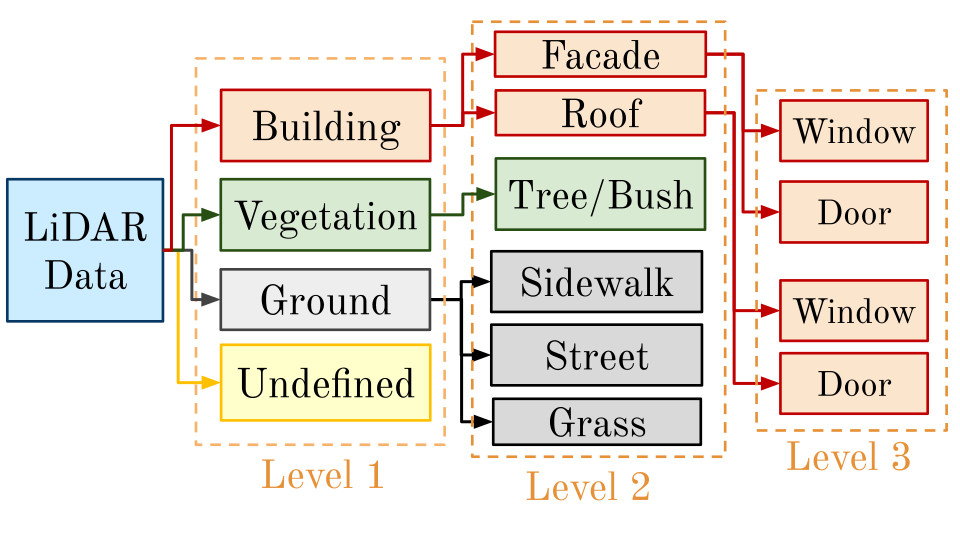}}\tabularnewline
(a) Overview of data and the selected area where labelled areas highlighted in red & (b) Hierarchy of classes (all 13 classes in 3 levels)\tabularnewline

 \end{tabular}
 }
 
{\caption{\small{Overview of the database}}
\label{fig:labels}}
\vspace{-1.8em}
\end{figure}

\textbf{LiDAR data}. The LiDAR point cloud used in this paper is derived from a registered point cloud of all these strips. Since the whole stacked point cloud includes more than 1.4 billion points, they are split into smaller tiles to be loaded and processed efficiently. The Dublin City airborne dataset is one of the world's densest urban ALS dataset ever collected. The final registered LiDAR point cloud offers an average density of 250 to 348 points/m${^2}$ in various tiles. Figure \ref{fig:labels}a shows an overview of the whole LiDAR point cloud that is colourised with regard to the elevation of the points. The selected area for labelling is highlighted inside the red boxes. 

In this paper, around 260 million points (out of 1.4 billion) were labelled. The selected area is within the most densely sampled part of the point cloud with full coverage by aerial images. This area (\ie inside the red boxes) includes diverse types of historic and modern urban elements. Types of buildings include offices, shops, libraries, and residential houses. Those buildings are in the form of detached, semi-detached and terraced houses and belong to different eras (\eg from 17${^{th}}$ century rubrics building to the 21${^{th}}$ century George's Quay complex as a modern structure).

In the initial data acquisition, generating LiDAR point cloud was the primary output of the aerial scanning. However, the helicopter also collected imagery data during the flight~\cite{laefer20172015}. Two different sets of images are detailed:

\textbf{Top view imagery}.\label{subsec:GeoRegisteredRGB} The dataset includes 4471 images taken from a helicopter and are named in the cited repository as Geo-referenced RGB. The resolution in pixels of the images is $9000\times6732$ and are stored in TIFF format, with a ground sampling distance of 3.4 cm. The total size of the images is around 813 GB and they are presented in the different flight paths. The geographic information is given as a GPS tag in the EXIF metadata and the camera used for the capture is Leica RCD30. 

\textbf{Oblique view imagery}.\label{subsec:ObliqueImagery} The oblique imagery contains 4033 JPEG images, which are taken with two different NIKON D800E cameras. They are referred to as Oblique photos, their resolution is $7360\times4912$ and its size is 18.5 GB. As per the geo-registered RGB images, they are presented in accordance with the flight path but include an extra subdivision which corresponds to each camera.

\textbf{Manual Labelling Process}. Herein, a manually annotated airborne LiDAR dataset of Dublin is presented. A subset of 260 million points, from the 1.4 billion laser point cloud obtained, have been manually labelled. The labels represent individual classes and they are included in three hierarchical levels (Figure \ref{fig:labels}b):
\begin{enumerate}[label=\roman*.]

  \item \textbf{Level 1:}  This level produces a coarse labelling that includes four classes: (a) Building; (b) Ground; (c) Vegetation; and (d) Undefined. Buildings are all shapes of habitable urban structures (\eg homes, offices, schools and libraries). Ground mostly contains points that are at the terrain elevation. The Vegetation class includes all types of separable plants. Finally, Undefined points are those of less interest to include as urban elements (\eg bins, decorative sculptures, cars, benches, poles, post boxes and non-static object). Approximately 10\% of the total points are labelled as undefined and they are mostly points of river, railways and construction sites. 
  \item \textbf{Level 2:}  In this level, the first three categories of Level 1 are divided into a series of refined classes. Buildings are labelled into roof and facade. Vegetation is divided into separate plants (\eg trees and bushes). Finally, Ground points are split into street, sidewalk and grass. 
  \item \textbf{Level 3:} Includes any types of doors and windows on roofs (\eg dormers and skylights) and facades.  
  
\end{enumerate}

\begin{figure}[!htb]
  \begin{center}
  \includegraphics[width=.95\textwidth]{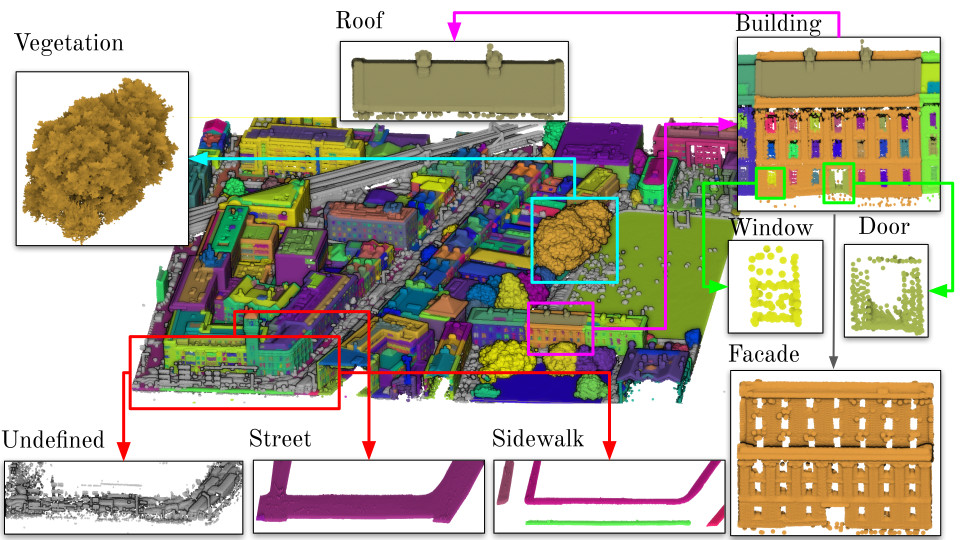}
  \caption{\small{A Sample labelled sub-tile that visualises all classes}}
  \label{fig:sceneClasses}
  \vspace{-1.8em}
  \end{center}
\end{figure}

In order to label the LiDAR data, it is divided into smaller sub-tiles (\ie each includes around 19 million laser scanning points). The process starts with importing data into the CloudCompare 2.10.1 \cite{cloudcompare} software. Then, points were coarsely manually segmented with segmentation and slicing tools in three categories (\ie building, vegetation and ground) and labelled accordingly. Then, the process continues to the third level which has the finest details (\ie windows and doors). Thereby, this pipeline produces a unique label for each point. The process is performed in over 2500 hours with an appropriate tutorial, supervision, and carefully crossed-checked multiple times to minimise the error. The naming order and the detail of labelling are demonstrated in more detail within the supplementary material. Figure \ref{fig:sceneClasses} shows the visual representation of the classes for one sub-tile which consists of around 19 million laser scanning points. 

%
\section{Results and Applications}
\begin{figure}
\footnotesize
\centering
 \resizebox{.9\textwidth}{!}{
 \def\arraystretch{1.5}
 \begin{tabular}{p{0.45\textwidth}p{0.45\textwidth}}
{\includegraphics[width=0.34\textwidth]{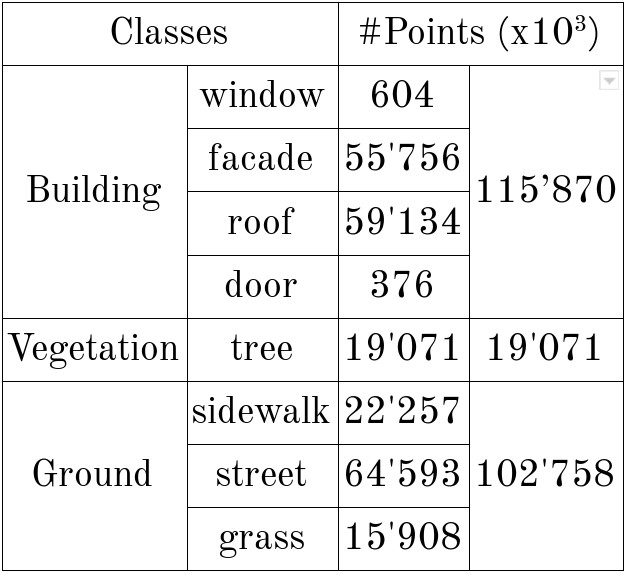}}&
{\includegraphics[width=.42\textwidth]{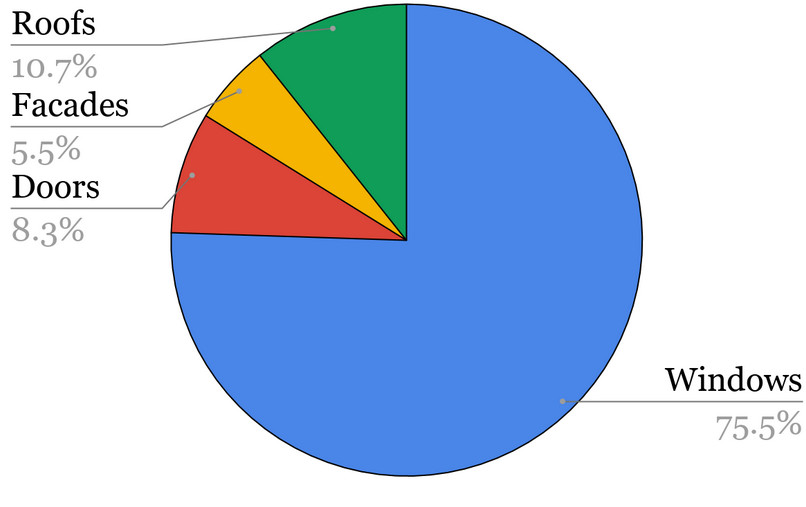}}\tabularnewline
(a) Total number of labelled points for each class & (b) Percentage of assets only in building class \tabularnewline
 \end{tabular}
}
{\caption{\small{Overview of the labelled results}}
\vspace{-1.6em}
\label{fig:pie}}
\end{figure}

As shown in Figure \ref{fig:pie}a, the majority of captured points belong to the ground classes. This is expected because point cloud density for the horizontal planes reflects more points towards the aerial scanner in comparison to the vertical surfaces (\eg facades). Also, windows have fewer points as laser beam usually penetrates the glass and there are only a few points on it. Similarly, the number of points in the door class is smaller as each building normally has one door for its entrance and one door for access to the roof. While this table shows the total number of points, Figure \ref{fig:pie}b shows the frequency percentage of each class that building category contains. For example, around 75\% of the objects in the building category are windows and around 8.3\% are doors as the number of windows in each building is much higher than other classes. 

The manually annotated point cloud dataset is available at \url{https://v-sense.scss.tcd.ie/DublinCity/}. An additional video is provided in the attached supplementary material for visual demonstration of the dataset. In the next sections, two applications are showcased by employing the annotated point cloud dataset.

\subsection{3D Point Cloud Object Classification}
The performance analysis of the dataset for object classification problem is showcased by using the state-of-the-art models namely, PointNet~\cite{qi2017pointnet}, PointNet++ ~\cite{qi2017pointnetplusplus} and SO-Net~\cite{li2018sonet}. These three models directly work on unstructured point cloud datasets. They learn the global point cloud features that have been shown to classify forty man-made objects of the ModelNet40~\cite{wu20153d} shape classification benchmark.

The 3D point cloud dataset is comprised of a variety of outdoor areas (\ie university campus and city centre) with structures of facades, roads, door, windows and trees as shown in Figure ~\ref{fig:sceneClasses}. In order to study the classification accuracy on the three CNN-based models, a dataset of $3982$ objects of $5$ classes (\ie doors, windows, facades, roofs and trees) is gathered. 

To evaluate the three models, the dataset is split into a ratio of $80:20$ for training and testing respectively. While training, for each sample, points on mesh faces are uniformly sampled according to the face area and normalised into a unit sphere (\ie -1 to +1). Additionally, data augmentation techniques are applied on-the-fly by randomly rotating the object along the up-axis and jittering the position of each point by Gaussian noise with zero mean and 0.02 standard deviation. Each model is trained for 100 epochs. 

In Table~\ref{tab:classification}, the performance of the three trained models in a different point cloud input setting using the Overall and Average class accuracy (as used in~\cite{qi2017pointnet,qi2017pointnetplusplus}) is shown. It is observed that with an increase in the number of points per objects, the performance of the three models increases. Amongst all the three networks, the So-Net architecture performs the best. This is in consistence with the results in~\cite{li2018sonet}. However, there is still a huge potential in the improvement of the performance scores. This is primarily because dataset is challenging in terms of structural similarity of outdoor objects in the point cloud space namely, facades, door and windows.

\begin{table}[h]
\begin{center}
\resizebox{.8\textwidth}{!}{
\def\arraystretch{1.5}
\begin{tabular}{|c|c|c|c|c|c|c|}
\cline{2-7} 
\multicolumn{1}{c}{} & \multicolumn{2}{|c|}{\textbf{PointNet}~\cite{qi2017pointnet} } & \multicolumn{2}{c|}{\textbf{PointNet++} ~\cite{qi2017pointnetplusplus}} & \multicolumn{2}{c|}{\textbf{So-Nets}~\cite{li2018sonet}} \tabularnewline
\hline 
 $\#$Points & Avg. Class & Overall & Avg. Class & Overall & Avg. Class & Overall\tabularnewline
\hline 
 512 & 24.17 & 35.17 & 39.47 & 45.56 & 41.89 & 48.74 \tabularnewline
 \hline
 1024 & 38.84 & 50.13 & 44.65 & 62.91  & 45.73 &  63.54\tabularnewline
 \hline
 2048 & 46.77 & 59.68 & 49.23 & 63.42 & 49.34 & 64.55 \tabularnewline
 \hline
 4096 & 48.77 &  60.68 & 51.23 & 64.42 & 50.34 & 65.55 \tabularnewline
 \hline
\end{tabular}
}
\end{center}
\caption{\small{Overall and Avg. class classification scores using the state-of-the-art models on the dataset.}}
\label{tab:classification}
\vspace{-1.6em}
\end{table}
\subsection{Image-based 3D Reconstruction}

In this section, the whole extension of the LiDAR point cloud described in Section~\ref{sec:Database} is exploited beyond the annotated data. To do that, an evaluation of the image-based 3D reconstruction is presented using two different types of aerial imagery data that are collected alongside the LiDAR data~\cite{laefer20172015}. The first set is composed of images with a top view of the city (by a downward-looking camera) and the second set consists of oblique views (by a $\approx 45^{\circ}$ tilted camera). More details of the images are given in Section~\ref{sec:Database}.  %

\begin{figure}
    \centering
    \includegraphics[height=4.7cm]{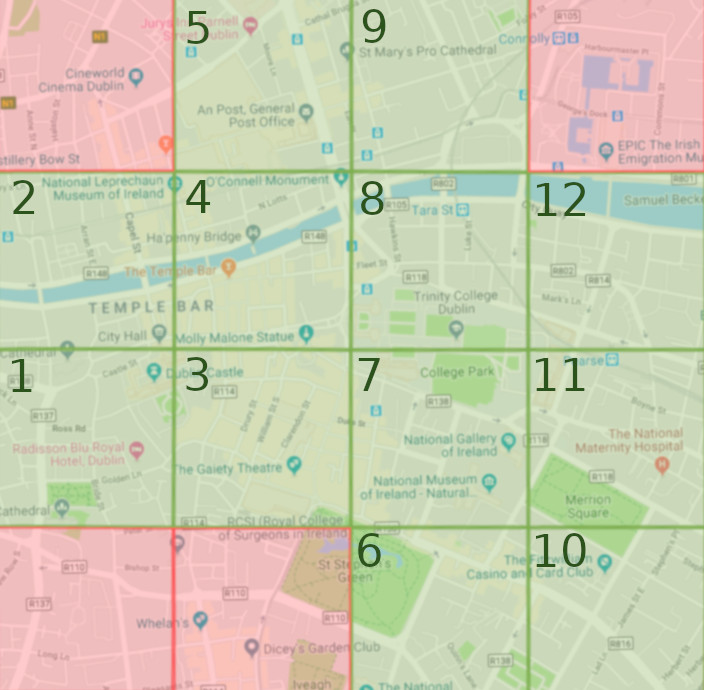}
     \includegraphics[height=5.1cm]{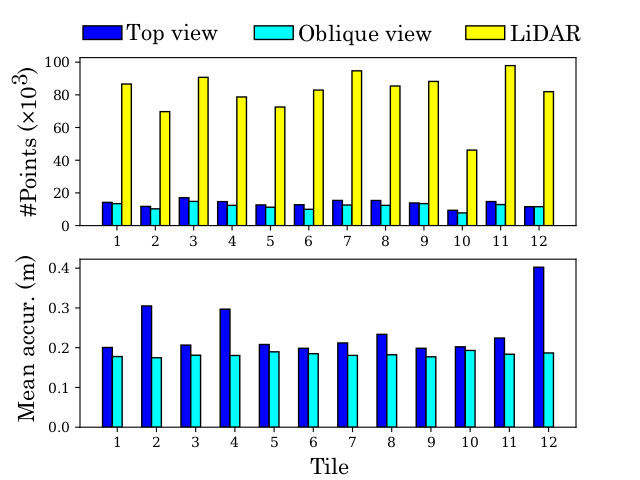}
    \caption{\label{fig:3dRecons:tilesAndAccuracy}\small{On the left, the area of the city covered by the tiles in the comparison (in green). On the right, a comparison of the number of points and the mean accuracy per tile.}}
    \vspace{-0.9em}
\end{figure}

In order to carry out the experiment, the complete reconstruction pipeline is evaluated as per~\cite{knapitsch2017tanks}. This is because the ground truth for the camera positions to specifically evaluate the camera poses from an SfM algorithm are not available, only the GPS position is known. The open-source software selected for the reconstruction is COLMAP (SfM~\cite{schoenberger2016sfm} and MVS~\cite{schoenberger2016mvs}), since it is reported as the most successful in different scenarios in the latest comparisons carried out (see Section~\ref{sec:RelatedWork}). Furthermore, it has advantages over other methods~\cite{moulon2013globalOPENMVG,barnes2009patchmatch} because it gives the possibility of handling a large amount of data without running out of memory. In this experiment, COLMAP is applied with the same configuration to each set of images and as a result, two dense point clouds are obtained. The configuration includes, apart from the default parameters, using a single camera per flight path and the vocabulary tree method~\cite{schoenberger2016vote} for feature matching. This was selected because it is the recommend mode for large image collections (several thousands). Moreover, as in COLMAP there is no option implemented to enforce GPS priors during SfM computation, we follow the recommendation of applying the geo-registration after obtaining the sparse point cloud. 

The point cloud associated with the top view images contains twenty-five million points whereas the one associated with the oblique images, containing twenty-two million points, is less dense. During the process, the point clouds are coarsely registered with the GPS information to the LiDAR point cloud. The GPS information available for the top view images is more accurate than the one available for the oblique ones but a fine registration with an ICP method~\cite{cloudcompare} (including scaling) is needed in both cases. The reconstructed point cloud is split into the same tiles as the LiDAR point cloud, each of which covers an area of 500x500 $m^ 2$. Overlap of these three point clouds is in twelve tiles only, which are shown in green in Figure~\ref{fig:3dRecons:tilesAndAccuracy}. This green area is the one under study, it is numbered for referential purposes and it allows for comparison of the reconstructions in different areas of the city.   

\begin{figure}[]
 \footnotesize
 \begin{tabular}{ccc}
 LiDAR & Top view reconst. (color) & Oblique view reconst. (color)\\
    \includegraphics[width=0.30\textwidth]{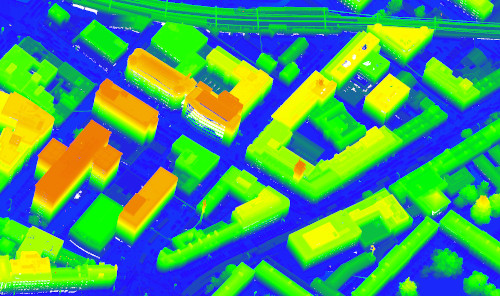} &
    \includegraphics[width=0.30\textwidth]{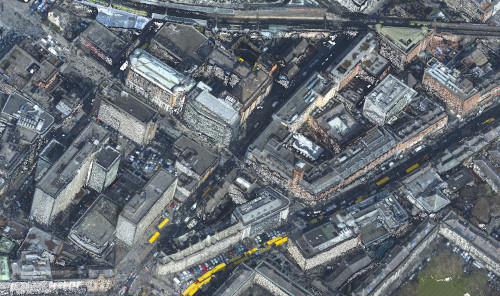} &
    \includegraphics[width=0.30\textwidth]{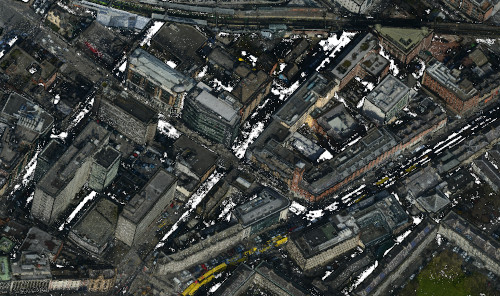}
\end{tabular}            
  \footnotesize
 \begin{tabular}{ccc}
 LiDAR & Top view reconst.  & Oblique view reconst.  \\
    \includegraphics[width=0.30\textwidth]{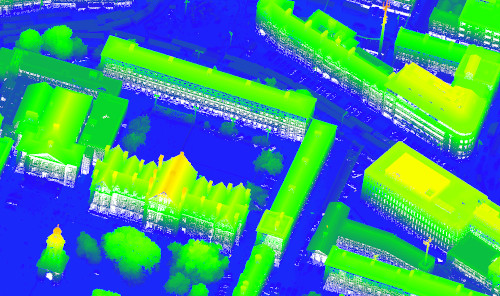} &
    \includegraphics[width=0.30\textwidth]{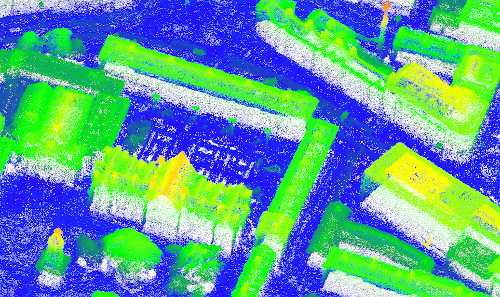} &
    \includegraphics[width=0.30\textwidth]{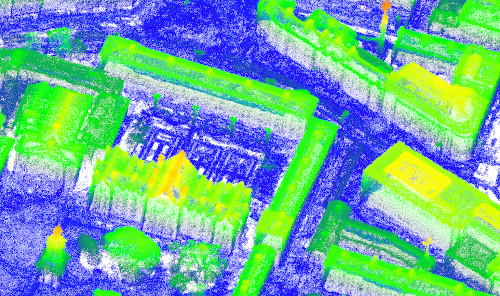}
\end{tabular}
 \caption{\label{fig:3dRecons:qualitativeResult}\small{\textbf{Qualitative results.} Comparison of the LiDAR with the image-based 3D reconstructions in two different parts of the city. Row I: including color. Row II: only geometry.}}
 \vspace{-1.5em}
\end{figure}

\begin{figure}
    \centering
    \includegraphics[width=0.32\linewidth]{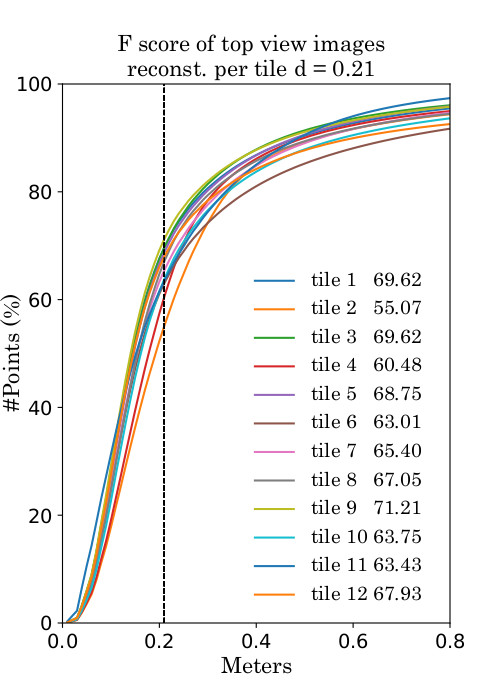}
     \includegraphics[width=0.32\linewidth]{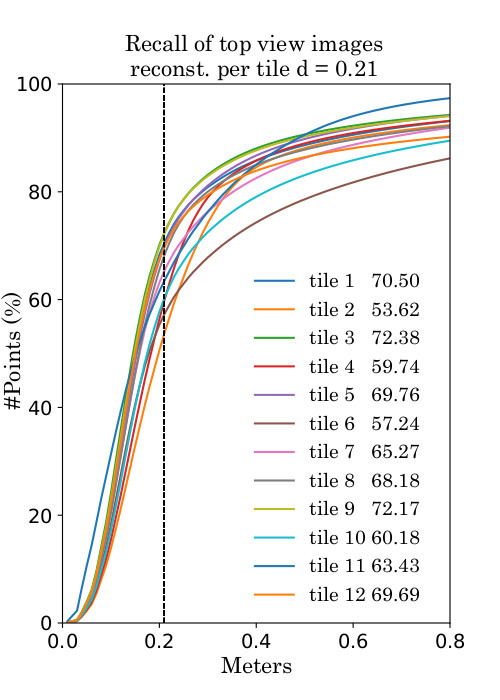}
     \includegraphics[width=0.32\linewidth]{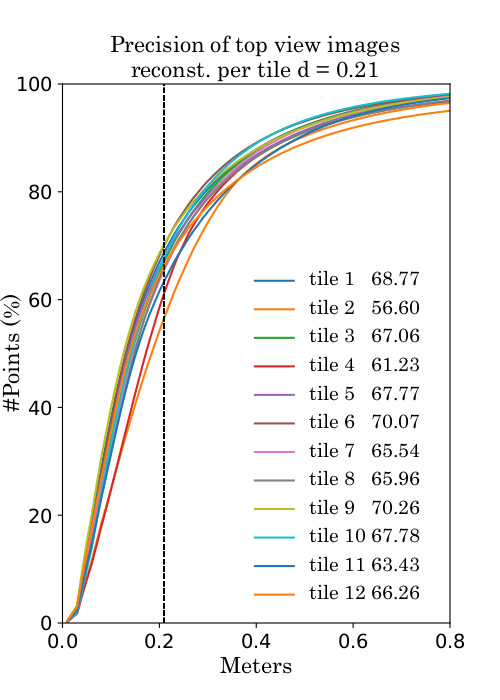}
     \vspace*{-5.5mm}
    \includegraphics[width=0.32\linewidth]{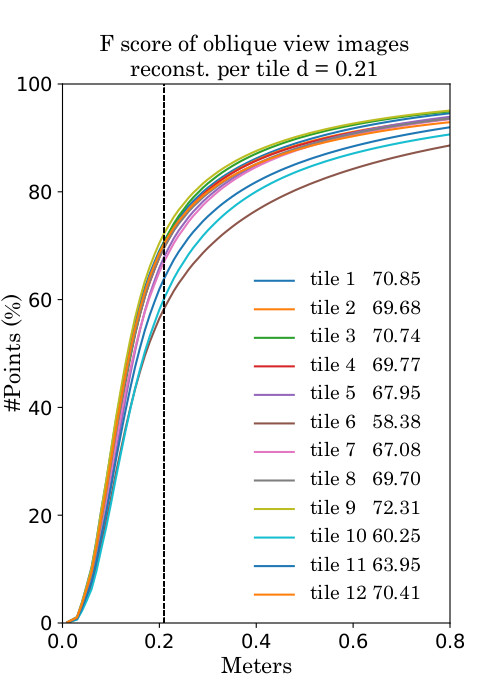}
     \includegraphics[width=0.32\linewidth]{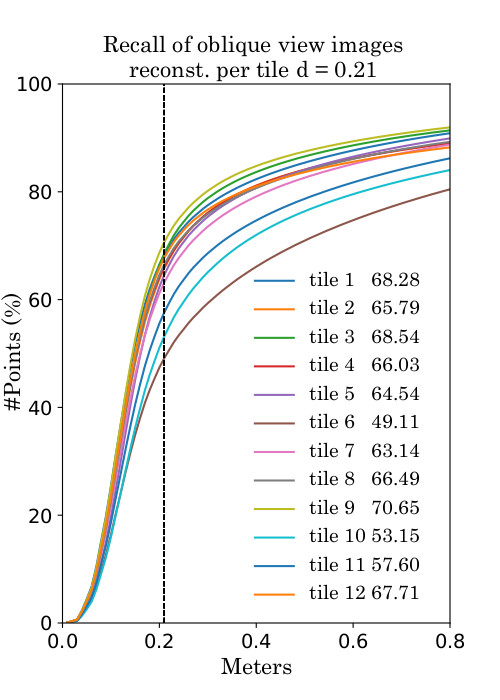}
     \includegraphics[width=0.32\linewidth]{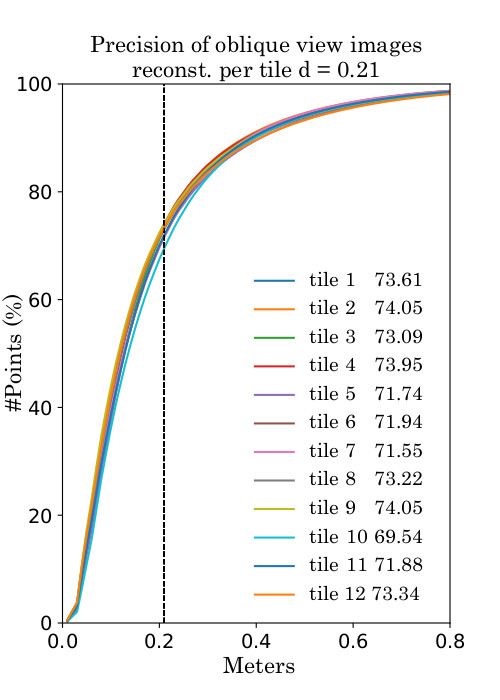}
    \caption{\label{fig:3dRecons:FscoreRecallPrecision}\small{F-score, recall and precision of the image-based reconstruction (from left to right) per tile. The top row shows the results of the top view images reconstruction and the bottom one, the results with the oblique view ones. The values given per tile corresponds with the value of the curves at a given distance (dash line).}}
    
\end{figure}

Some qualitative results are shown in Figure~\ref{fig:3dRecons:qualitativeResult} and from a quantitative perspective, it can be observed in Figure~\ref{fig:3dRecons:tilesAndAccuracy} that the LiDAR point cloud is more than four times denser than the ones obtained with image-based reconstruction. A denser image-based reconstruction is obtained with the top view data in every tile. At the bottom, the same figure shows the mean accuracy from each reconstruction to the LiDAR point cloud. It shows that, on average, the oblique imagery is closer to the ground truth. This difference is approximately 2 cm in the majority of them, however, 3 of the tiles (2, 4 and 12) present a larger difference. As can be seen in Figure~\ref{fig:3dRecons:tilesAndAccuracy}, these correspond to the tiles with a river, and they are followed by tile 8, which is also in the area. %


As pointed out in~\cite{knapitsch2017tanks}, the mean distance between the point clouds can be affected by outliers. Hence, they propose to use the following measurements for further study: precision, recall, and F score. The precision, $P$, shows the accuracy of the reconstruction, the recall,  $R$, is related to how complete the reconstruction is, and the F score, $F$, is a combination of both. 
They are defined in Eq.~\eqref{eq:1} for a given threshold distance $d$. In the equations, $I$ is the image-based reconstruction point cloud, $G$ is the ground truth point cloud, $ | \cdot| $ is the cardinality, $dist_{I \rightarrow G }(d)$ are the points in $I$ with a distance to $G$ less than $d$ and $dist_{G \rightarrow I }(d)$ is analogous (\ie $dist_{A \rightarrow B}(d)= \{ a \in A \mid \min\limits_{b \in B} \Arrowvert a - b \Arrowvert_{2} < d \} $, $A$ and $B$ being point clouds).  

\begin{align}
    P(d) &= \frac{| dist_{I \rightarrow G }(d)|}{|I|} 100 &
    R(d) &= \frac{| dist_{G \rightarrow I }(d)|}{|G|}100 &
    F(d) &= \frac{2P(d)R(d)}{P(d)+R(d)}
    \label{eq:1}
\end{align}

The results of these three measurements are given in Figure~\ref{fig:3dRecons:FscoreRecallPrecision}. From these, the results of the top view reconstruction are more tile dependant than the ones obtained with the oblique imagery. The precision in the latter is very similar for every tile, however, the recall is much lower in tiles 6, 10 and 11. A commonality amongst these tiles is that they contain part of the parks of the city, as shown in Figure~\ref{fig:3dRecons:tilesAndAccuracy}. Apart from that, tile 9 has a higher F score, which corresponds to an area without any river or green areas.

    

%
\section{Conclusions and Future Work}

This paper presents a highly dense, precise and diverse labelled point cloud. Herein, an extensive manually annotated point cloud dataset is introduced for Dublin City. This work processes a LiDAR dataset that was unstructured point cloud of Dublin City Centre with various types of urban elements. The proposed benchmark point cloud dataset is manually labelled with over 260 million points comprising of 100'000 objects in 13 hierarchical multi-level classes with an average density of $348.43$ points/m$^2$. The intensive process of labelling is precisely cross-checked with expert supervision. The performance of the proposed dataset is validated on two salient applications. Firstly, the labelled point cloud is employed for classifying 3D objects using state-of-the-art CNN based models. This task is a vital step in a scene understanding pipeline (\eg urban management). Finally, the dataset is also utilised as a detailed ground truth for evaluation of image-based 3D reconstructions. The dataset will be publicly available to the community.

In addition to the above, the usage of the dataset can be extended in several applications. For example, the annotated dataset can be used for a further evaluation of image-based 3D reconstruction per class instead of per tile. Also, it can be employed for object segmentation in remote sensing or Geographic Information System (GIS), volumetric change detection for forest monitoring, as well as in disaster management. Additionally, it can be applied to optimise traffic flow for smart cities and even for the generation of real models of large cityscapes in the entertainment industry.  


\section{Acknowledgement}
\footnotesize
This publication has emanated from research supported in part by a research grant from Science Foundation Ireland (SFI) under the Grant Number 15/RP/2776 and in part by the European Union's Horizon 2020 Research and Innovation Programme under Grant Agreement No 780470. 
The authors highly appreciate the original work of generating LiDAR Point Cloud at Urban Modelling Group in University College Dublin in 2015. In addition, we are grateful for all of the volunteers who generously participated in the process of data labelling, especially Mr S Pouria Vakhshouri Kouhi for his constant support. We also gratefully acknowledge the support of NVIDIA Corporation with the donation of the Titan Xp GPU used for this research.

\small
\bibliographystyle{ieee}
\bibliography{ms}

\end{document}